\pdfoutput=1

\documentclass[11pt]{article}

\usepackage[preprint]{acl}

\usepackage{times}
\usepackage{latexsym}

\usepackage[T1]{fontenc}

\usepackage[utf8]{inputenc}

\usepackage{microtype}

\usepackage{inconsolata}

\usepackage{graphicx}
\usepackage{subfigure}
\usepackage{multirow}
\usepackage{multicol}
\usepackage{booktabs}
\usepackage{amssymb}
\usepackage{bbding}
\usepackage{amsmath}
\usepackage{amsfonts}
\usepackage{amssymb}
\usepackage{hyperref}
\usepackage{threeparttable}
\usepackage{pifont}
\usepackage{listings}

\title{BLSP: Bootstrapping Language-Speech Pre-training via Behavior Alignment}

\author{Chen Wang\textsuperscript{1,3}\thanks{Work was done while at Alibaba Inc.}, Minpeng Liao\textsuperscript{2}, Zhongqiang Huang\textsuperscript{2}\thanks{Corresponding author.}, Jinliang Lu\textsuperscript{1,3}, Junhong Wu\textsuperscript{1,3} \\
{\bf Yuchen Liu\textsuperscript{1}, Chengqing Zong\textsuperscript{1,3}, Jiajun Zhang\textsuperscript{1,3,4}\footnotemark[2]}
\\
\textsuperscript{1} Institute of Automation, Chinese Academy of Sciences\\
\textsuperscript{2} Machine Intelligence Technology Lab, Alibaba Inc.\\
\textsuperscript{3} School of Artificial Intelligence, University of Chinese Academy of Sciences \\
\textsuperscript{4} Wuhan AI Research \\
\texttt{\{wangchen2020\}@ia.ac.cn} \texttt{\{jjzhang,cqzong\}@nlpr.ia.ac.cn}\\
\texttt{\{minpeng.lmp,z.huang\}@alibaba-inc.com} \\
}

\begin{document}
\maketitle
\begin{abstract}
The emergence of large language models (LLMs) has sparked significant interest in extending their remarkable language capabilities to speech. However, modality alignment between speech and text remains an open problem. Current solutions can be categorized into cascaded approaches, which limit the interaction between speech and LLMs, and end-to-end approaches that rely on scarce speech instruction data. In this paper, we propose the \textbf{BLSP} approach that \textbf{B}ootstraps \textbf{L}anguage-\textbf{S}peech \textbf{P}re-training via behavior alignment, leveraging existing ASR training data. We achieve this by developing a lightweight modality adapter between a frozen speech encoder and an LLM, optimized to ensure that the LLM exhibits the same generation behavior irrespective of the modality of input: a speech segment or its transcript. We primarily focus on the continuation writing behavior as it closely resembles next-token prediction in a broad sense but also found that introducing other behaviors could lead to improved performance. We demonstrate that this simple process can extend the capabilities of LLMs to speech and achieve competitive performance compared to cascaded systems, enabling speech recognition, speech translation, spoken language understanding, and speech conversation, even in zero-shot cross-lingual scenarios.\footnote{Please find code and model at \url{https://github.com/cwang621/blsp}. Video demos are available at \url{https://cwang621.github.io/blsp.github.io/}.}

\end{abstract}

\section{Introduction}

Large Language Models (LLMs), trained on massive amounts of textual data, have achieved significant success on various natural language processing tasks \citep{chowdhery2022palm,openai2023gpt,gao2023llama}. Recent research efforts have attempted to expand LLMs' capabilities to comprehend diverse modalities \citep{yin2023survey,latif2023sparks}. Speech, as an important modality, offers a plethora of benefits that complement text-based communication. Speech not only serves as the primary mode of human interaction but also conveys rich emotions, tones, and intentions that cannot be fully captured in text. Thus, enabling LLMs to understand speech could greatly enhance their utility in real-world scenarios.

However, effectively integrating and aligning speech with LLMs remains a significant challenge. Current approaches can be classified into two categories. One adopts a cascade paradigm, where the LLM is equipped with an automatic speech recognition (ASR) model to convert speech into text \citep{huang2023audiogpt,shen2023hugginggpt}, or the LLM is fed output states from a separately trained recognition system \citep{chen2023x}. In this setup, the transfer of knowledge from the LLM to the speech modality is hindered due to the separation between ASR and LLM training. Recent efforts explore training end-to-end speech-language LLMs for direct speech interaction \citep{zhang2023speechgpt,shu2023llasm}. Yet, this approach heavily relies on scarce speech instruction data, which is challenging to collect in large quantities, and struggles to generalize robustly across languages and speakers. In this work, we address the question of whether it is possible to align speech and text in a generalized manner using existing cross-modal datasets like ASR data, which is available in large volumes.

Our preliminary investigation has revealed that a model trained to predict the ground-truth transcript with speech input loses the ability to follow instructions. To achieve effective cross-modal alignment, we introduce the BLSP approach, which bootstraps language-speech pre-training via behavior alignment. The key idea is to develop a lightweight modality adapter between a frozen speech encoder and an LLM, optimized to ensure that the LLM exhibits the same generation behavior irrespective of the modality of input: a speech segment or its transcript. Specifically, we first prompt an LLM to generate text responses from speech transcripts. Then, we use these responses as supervised signals to optimize the parameters of the modality adapter. Our primary focus is on the continuation writing behavior as it prompts the LLM to generate text that resembles the broad data it is trained on, without biasing toward a specific task. However, we have observed that incorporating other behaviors, specifically repetition that mirrors the speech recognition task, could yield advantages in fine-grained lexical modeling. Our experiments reveal that the BLSP approach can effectively achieve cross-modal alignment and achieve competitive performance compared to cascaded systems, enabling LLMs to understand speech while retaining their language capabilities.


The contributions of our work are as follows:
\begin{itemize}
    \item We introduce a novel approach to bootstrap language-speech pre-training through behavior alignment, providing a new direction for cross-modal alignment in LLMs.

    \item We develop a simple process that requires training only a lightweight modality adapter, leveraging a pretrained speech encoder and LLM, and utilizing existing speech recognition data, thus eliminating the need to acquire speech instruction data.

    \item We conduct quantitative evaluations and provide video demonstrations to showcase that our BLSP approach effectively extends LLMs to speech inputs and achieves competitive performance compared to cascaded systems, enabling speech recognition, speech translation, spoken language understanding, and speech conversation, even in zero-shot cross-lingual scenarios.
\end{itemize}

\section{Background} \label{sec:background}

Due to the scarcity of speech instruction data, a natural approach to align speech and text for leveraging LLMs is to connect a pre-trained speech encoder to an LLM through a modality adapter trained on large volumes of speech-transcript pairs collected for speech recognition, as explored in \cite{shu2023llasm,xue2024echat}.
Similar methods have achieved considerable success in vision-language models. Notably, BLIP-2 \citep{li2023blip} and MiniGPT-4 \citep{zhu2023minigpt} have demonstrated that training a learnable interface using aligned image caption data can effectively bridge the modality gap between vision and language, enabling an LLM to comprehend images while retaining its capacity to follow text prompts.

\begin{figure}
    \centering
    \includegraphics[width=0.5\textwidth]{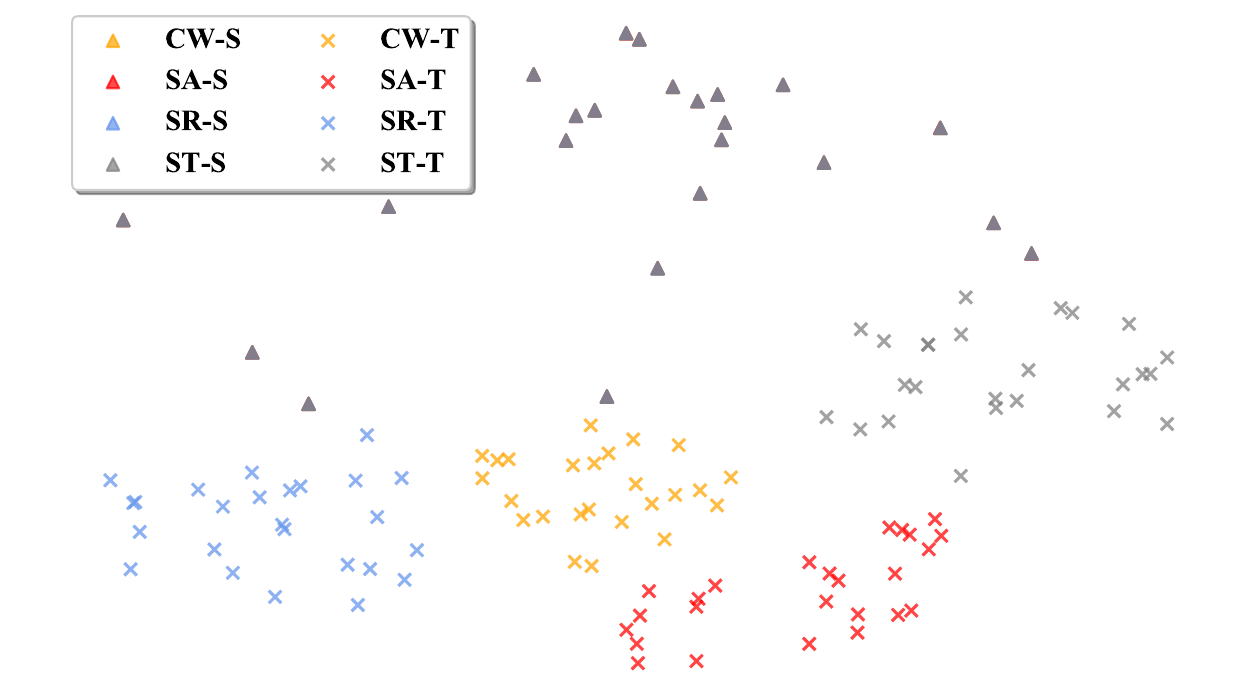}
    \caption{T-SNE visualization of feature representations learned from ASR task. Colors denote task instructions: orange for continuation writing (CW), red for sentiment analysis (SA), blue for speech recognition (SR), and gray for speech translation (ST). Shapes distinguish input modality: triangles for speech, crosses for text. Note that speech inputs result in overlapping features.}
    \label{fig:vis_asr}
\end{figure}

However, this approach proves to be more intricate when it comes to effectively achieving speech and text alignment, crucial for extending the language capabilities of LLMs to speech inputs. Our preliminary investigation, detailed in Appendix \ref{app:background}, has found that training a modality adapter to predict the ground-truth transcript from speech input can inadvertently restrict the LLM to performing solely speech recognition tasks. This issue arises despite the variety of transcription instructions used during training, as the LLM tends to overlook any textual instructions provided before the speech segment. We hypothesize that the reliance on homogeneous ASR training data results in a strong bias in the learned speech representations, confining the LLM's functionality to the transcription task only.


\begin{figure*}[tp]
\centering
\includegraphics[width=0.9\textwidth]{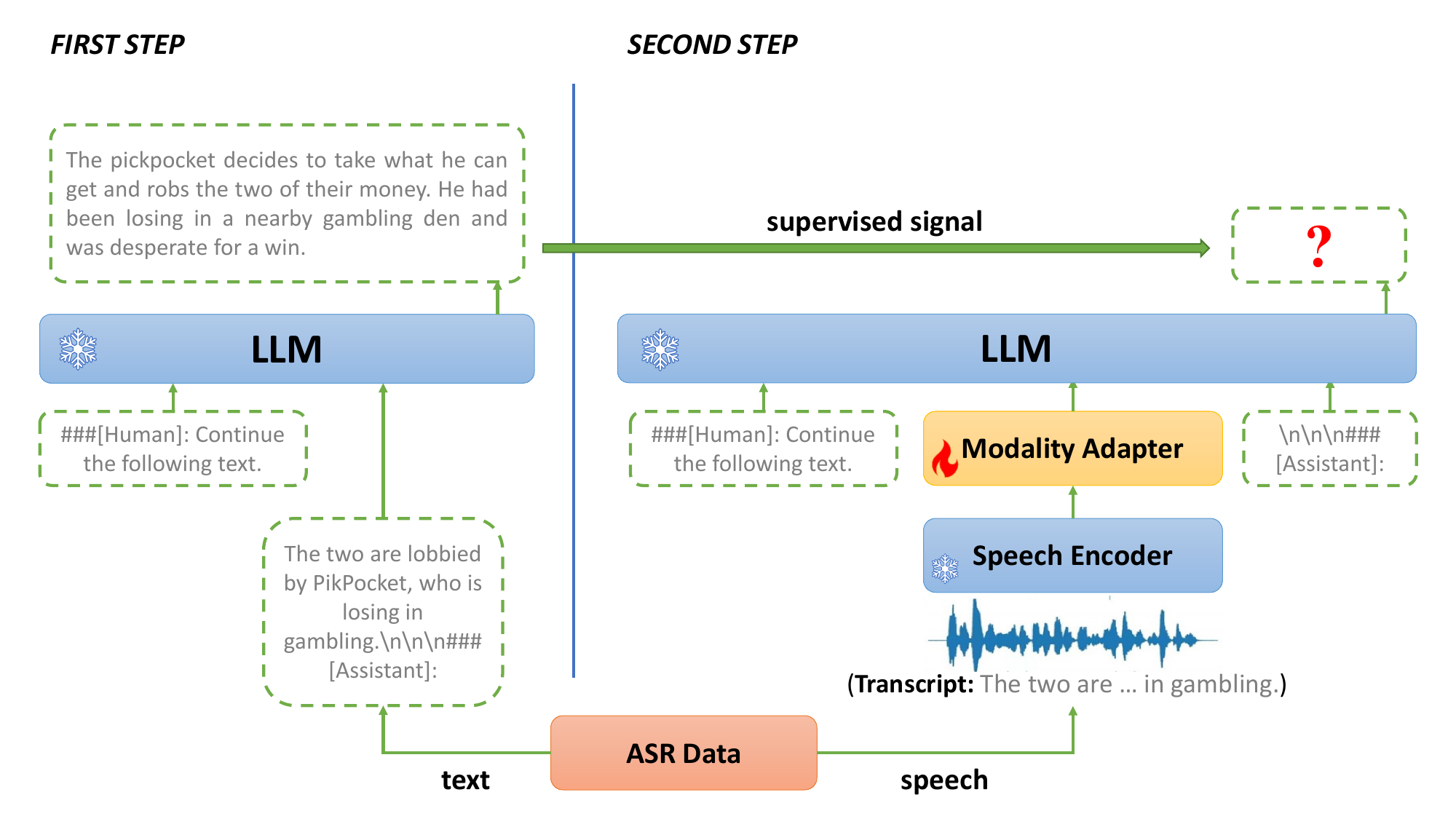}
\caption{An overview of our BLSP approach for behavior alignment. Text response generated given speech transcripts as inputs by an LLM (in the first step) are used as supervisions to train the modality adapter given the corresponding speech as inputs (in the second step).}  
\label{fig:cmba}
\end{figure*}

To substantiate our hypothesis, we conducted an analysis of the representations learned from ASR task on speech and text pairs from the LibriSpeech dataset \citep{panayotov2015librispeech}. We consider four distinct tasks: continuation writing (CW), sentiment analysis (SA), speech recognition (SR), and speech translation (ST), prompt by instructions. For each task, the same task-specific instruction is employed to prompt the LLM to process either speech or its corresponding transcript. The cross-modal prompt is formatted as follows:

\begin{small}
\begin{verbatim}
###[Human]:<instruction><speech/transcript>\n\n\n
###[Assistant]:
\end{verbatim}
\end{small}

The learned representations of these inputs are obtained by extracting the hidden state of the final layer for the last token of the cross-modal prompt, right before response generation. We provide a visualization of the representations of 25 samples in Figure \ref{fig:vis_asr}. Ideally, paired speech and transcript inputs should yield similar representations, and these representations should be clustered based on task instructions. However, this visualization clearly demonstrates the separation between speech and text representations in the feature space. The LLM consistently projects speech input into almost identical representations regardless of the instructions provided, resulting in overlapping markings in the figure. This indicates a lack of ability to adhere to instructions when processing speech inputs. We provide quantitative evidence in Appendix \ref{app:background}. This inability to bridge the modality gap through ASR tasks prompts us to reevaluate what it means to align speech and text for LLMs.

\section{Method}

Our proposed approach, named \textbf{B}ootstrapping \textbf{L}anguage-\textbf{S}peech \textbf{P}retraining (\textbf{BLSP}) via behavior alignment, is designed to align pre-trained speech encoders with large language models (LLMs), with the goal of extending the remarkable language capabilities to speech input. Our model comprises three components: a speech encoder, an instruction-following LLM, and a modality adapter between the speech encoder and LLM. We keep both the speech encoder and the LLM frozen during the training process and only train the parameters of the modality adapter. An overview of our model is presented in Figure \ref{fig:cmba}. We will next describe how to construct data to train the modality adapter in an end-to-end manner.


Instead of treating the speech-transcript pairs as input-output mappings, we consider the speech and its transcript in each pair as two independent inputs to the LLMs. Intuitively, if the representations of a speech segment are well aligned in the textual space for an LLM, the LLM should behave the same no matter whether it is given the speech segment or its transcript as input, under the same instruction. In other words, it should generate the same text. We term this concept \emph{behavior alignment}. 

Based on this concept, the BLSP approach consists of two steps. In the first step, we use speech transcripts as input and instruct the LLM to generate responses using the following prompt:

\begin{small}
\begin{verbatim}
###[Human]:<instruction><transcript>\n\n\n
###[Assistant]:
\end{verbatim}
\end{small}

In the second step, we feed the corresponding speech signals to the speech encoder and use the speech representations produced by the modality adapter as the input to the LLM, replacing the word embeddings of the transcripts. We regard the responses produced in the first step as the ground truth for supervised learning using language modeling loss with the following prompt:

\begin{small}
\begin{verbatim}
###[Human]:<instruction><speech>\n\n\n
###[Assistant]:<response>
\end{verbatim}
\end{small}

In this work, we primarily focus on the continuation behavior, as it resembles universal generation in next-token prediction and can produce a diverse range of texts reflecting the extensive dataset used to train LLMs. This characteristic is crucial as it avoids over-fitting to more specific behaviors that exhibit a strong bias in the response, akin to issues encountered with ASR pretraining, as discussed in Section \ref{sec:background}.

While not inherently beneficial on their own, incorporating data from certain specific behaviors alongside continuation data in modest proportions can enhance the performance of the model. For instance, integrating the repetition behavior, which resembles ASR pretraining, into the continuation data can assist the adapter in capturing fine-grained lexical information, as explored in this study. We leave the systematic investigation of other behaviors for future research. 

See Table~\ref{tab:ins_method} for two instructions used to prompt behaviors. It is worth noting that since the response for repetition behavior is simply the original transcript with minor changes based on how closely the LLM follows the repetition instruction, we can skip the first step and directly use the speech transcript as the response.


\begin{table}[htbp]
    \centering
    \footnotesize
    \begin{tabular}{p{0.45\textwidth}}
    \toprule
    \textbf{Continuation:} \emph{Continue the following text in a coherent and engaging style with less than 40 words.} \\
    \midrule
    \textbf{Repetition:} \emph{Please repeat the following words.} \\
    \bottomrule
    \end{tabular}
\caption{Instructions used to prompt LLM behaviors.}
\label{tab:ins_method}
\end{table}

\section{Experiment Setup}
\subsection{Training Details}
\label{sec:training_details}

We utilize the encoder part of Whisper-small \citep{radford2022robust} as the speech encoder and employ Llama-2-7B \citep{touvron2023llama} as the large language model (LLM). To induce instruction-following capability\footnote{We could have also used the official chat version of Llama-2, but we opted to perform instruction finetuning using publicly available data, as it offers flexibility for future research involving multi-modal instruction data.}, we employ the publicly accessible dataset Alpaca-52K \citep{alpaca} to fine-tune the LLM. The Alpaca-52k dataset consists of 52K (text instruction, text input, response) triplets, which we convert into (text instruction, response) pairs by combining the instructions and inputs. During this stage, we fine-tune all parameters of the LLM for 3 epochs with a batch size of 128. 

The modality adapter is composed of three 1-dimensional convolution layers followed by a bottleneck layer \citep{houlsby2019parameter} with a hidden dimension of 512. The convolution layers are designed to reduce the length of the speech features by a factor of 8, with each layer having a stride size of 2, a kernel size of 5, and a padding of 2. To train the modality adapter, we utilize publicly available speech recognition datasets, including LibriSpeech \citep{panayotov2015librispeech}, GigaSpeech \citep{chen2021gigaspeech}, and Common Voice 2.0 \citep{commonvoice:2020}.

We train two BLSP models. The primary BLSP model is trained solely on continuation behavior, using 8.8 million (speech, text continuation) pairs obtained by performing the continuation writing task on the ASR datasets with the fine-tuned Llama-2 model. The secondary BLSP+RP model, included for comparison, is trained on a 1:9 mixing ratio of repetition data in the form of (speech, transcript) pairs and the aforementioned continuation data. During this stage, we fine-tune the modality adapter for one epoch with a batch size of 768.


\subsection{Baselines}
\label{sec:baselines}

We compare our method with the following baselines.

\paragraph{Text+LLM} The input to the LLM is the ground-truth speech transcripts.

\paragraph{Whisper+LLM}  The input to the LLM is the speech recognition output from whisper-small, which is comprised of both an encoder (utilized as the speech encoder in BLSP) and a decoder (not employed in BLSP). When comparing BLSP models to this baseline, it is important to note that BLSP's speech training data is much smaller than that for Whisper models.

\paragraph{CTC+LLM}  The input to the LLM is the speech recognition output from an in-house CTC-based ASR model. This ASR model consists of a speech encoder and adapter identical to those in BLSP, in addition to a CTC projector. We freeze the speech encoder and fine-tune the adapter and projector on the same ASR datasets used for BLSP training. We consider CTC+LLM as the most realistic cascaded baseline for demonstrating the modeling power of the BLSP approach.

\begin{table*}[htbp]
\centering
\footnotesize
\begin{tabular}{lr|lr|lcccccc}
    \toprule
    \multirow{2}{*}{\textbf{Method}} & \multicolumn{4}{c}{\textbf{ASR (WER$\downarrow$)}} & \multicolumn{2}{c}{\textbf{ST (BLEU$\uparrow$)}} & \multicolumn{3}{c}{\textbf{SLU (ACC$\uparrow$)}}\\ 
    \cmidrule(lr){2-5} \cmidrule(lr){6-7} \cmidrule(lr){8-10}
     & \multicolumn{2}{c}{\textbf{LibriSpeech}} & \multicolumn{2}{c}{\textbf{TED-LIUM}} & \multirow{2}{*}{\textbf{MUST-C}} & \textbf{CoVoST} & \textbf{SNIPS} & \multirow{2}{*}{\textbf{FSC}} & \textbf{SLUE}\\
     & \multicolumn{2}{c}{\textbf{test-clean}} & \multicolumn{2}{c}{\textbf{3}} & & \textbf{2.0} & \textbf{light-close} & & \textbf{VoxCeleb}\\ 
     \toprule
    \textbf{Text+LLM} &  \phantom{0}0.0 & \phantom{0}5.6  & \phantom{0}0.0 & 14.5 & 19.7 & 21.9 & 86.3 & 72.4 & 75.0 \\
    \textbf{Whisper+LLM}  & \phantom{0}3.4 & \phantom{0}5.9 & \phantom{0}4.3 & 20.4 & 16.6 & 16.9 & 83.2 & 56.3 & 74.1  \\
    \textbf{CTC+LLM} & \phantom{0}6.2 & 10.8 & \phantom{0}8.4 & 20.7 & 13.3 & 13.2 & 79.0 & 60.4 & 74.7 \\
    \textbf{ASR pretraining} & — & \phantom{0}3.7 & — & \phantom{0}4.5 & \phantom{0}0.0 & \phantom{0}0.0 & \phantom{0}0.0  & \phantom{0}0.0 & \phantom{0}0.0 \\
    \midrule
    \textbf{BLSP} & — & 10.4 & — & 23.1 & 12.3 & 12.7 & 75.8 & 60.9 & 76.0 \\
    \quad +RP & — & \phantom{0}6.4 & — & \phantom{0}8.1 & 14.9 & 13.8 & 78.8 & 77.5 & 75.5 \\
    \bottomrule
\end{tabular}
\caption{Overview of BLSP results on zero-shot speech-to-text tasks. For each ASR test set, we report two WER scores: on the left for the standalone ASR component of a cascaded system, and on the right for instructing the LLM to repeat the words recognized by the ASR component. The BLEU scores for the ST test sets are averaged across multiple translation directions.}
\label{tab:zero_shot}
\end{table*}

\paragraph{ASR pretraining} The model utilizes the same architecture as BLSP, except the modality adapter is trained by predicting the ground-truth transcript, as discussed in Section \ref{sec:background} and detailed in Appendix \ref{app:background}.






\section{Results}
We have found through experiments that the proposed BLSP approach is capable of empowering the LLM with speech understanding capabilities while maintaining fidelity to text instructions, achieving competitive performance compared to the cascaded baseline CTC+LLM. We conduct evaluations on multiple speech-related downstream tasks, including speech recognition (ASR), speech translation (ST), and spoken language understanding (SLU). It is important to note that the primary BLSP model is trained solely on the continuation writing task; therefore, all evaluations are conducted in a \textbf{zero-shot} setting, where we utilize text instructions to perform various speech-to-text generation tasks. For the BLSP+RP model, all evaluations except the ASR task are conducted in a zero-shot setting. We also demonstrate the open-ended generation capability of BLSP by conducting general-purpose QA. Additionally, we demonstrate that our model supports cross-modal conversations and develops multilingual capabilities, even though the alignment training is carried out only in English.



\subsection{Quantitative Evaluations}

\paragraph{Instructed Zero-Shot Speech-to-Text Tasks}
We perform speech-to-text tasks by prompting the BLSP model with task-specific instructions, detailed in Table~\ref{tab:ins} in the Appendix, followed by the speech features as input to the LLM. The same instructions are also used in the baseline systems.


For the ASR task, we conduct quantitative evaluations on both in-domain (LibriSpeech, \citealp{panayotov2015librispeech}) and out-of-domain (TED-LIUM 3, \citealp{hernandez2018ted}) test sets, utilizing Word Error Rate (WER) as the evaluation metric. In a cascaded system, the ASR task can be performed in two ways: either directly using the standalone ASR component or by instructing the LLM to repeat the words recognized by the ASR component. We compare these methods to assess the LLM's ability to follow ASR instructions.
For the ST task, we use SacreBLEU \citep{post-2018-call} as the evaluation metric, and report in-domain results on CoVoST-2 \citep{wang2020covost} and out-of-domain results on MUST-C \citep{di2019must}, averaged across five and eight translation directions, respectively, as detailed in Appendix \ref{app:results}. 
For the SLU task, we evaluate on intent classification (IC) datasets SNIPS \citep{saade2019spoken} and FSC \citep{lugosch2019speech}, and sentiment analysis (SA) dataset SLUE-VoxCeleb \citep{shon2022slue}, using accuracy as the evaluation metric.

Results are presented in Table \ref{tab:zero_shot}. Despite a significant performance disparity in ASR and ST tasks when compared to the cascaded system of Whisper+LLM, our primary BLSP model demonstrates respectable outcomes across all evaluated tasks. It's important to note that the Whisper model, incorporating a decoder absent in our BLSP model, benefits from training on a substantially larger corpus of speech data than the BLSP model. However, when contrasted with the most directly comparable cascaded baseline, CTC+LLM, which has a similar architecture and is trained on an equivalent volume of speech data, the performance gap narrows considerably. Remarkably, the BLSP model surpasses CTC+LLM in the FSC and SLUE-VoxCeleb test sets for the SLU tasks. Conversely, the ASR pretraining method, frequently employed in prior research to facilitate cross-modal alignment in LLMs \citep{zhang2023speechgpt, shu2023llasm}, proves ineffective in maintaining any capability for instruction-following in non-ASR tasks.


\paragraph{Incorporating Additional Behaviors}
We observe that the ASR component of a cascaded system has a significantly lower WER score than the cascaded system itself. This suggests that the LLM's insufficient ability to closely follow the ASR instruction is one of the reasons the BLSP method performs less effectively than traditional ASR models in ASR tasks. As shown in Table \ref{tab:zero_shot}, the BLSP+RP model, which utilizes repetition training data at a 1:9 mixing ratio with the continuation training data, achieves WER scores comparable to the ASR component of the CTC+LLM model, and significantly better scores than those achieved by the CTC+LLM method through prompting. Moreover, the inclusion of repetition data also leads to improved performance on other tasks, achieving better scores on both ST test sets and two of the three SLU test sets compared to the CTC+LLM baseline.


\paragraph{General-Purpose QA}

We also evaluated the performance of our BLSP models on a general-purpose question-answering (QA) task. This task focuses on grasping the semantics conveyed through speech and encompasses a broader range of textual instructions. For this evaluation, detailed in Appendix \ref{app:qa}, we selected 1460 samples from the GigaSpeech test set and employed ChatGPT to create a question for each sample based on its transcript. We then utilized ChatGPT again to determine the acceptability of the responses generated by different methods. The evaluation findings are summarized in Table \ref{tab:qa}. Both BLSP models exhibited competence in this task, achieving scores comparable to the cascaded baseline CTC+LLM (88.5\%/88.3\% vs 88.6\%). This performance underscores our approach's ability to endow the LLM with a general comprehension of speech, thereby equipping it to adeptly handle diverse cross-modal instructions and produce satisfactory responses.

\begin{table}[tbp]
    \centering
    \begin{tabular}{l|c}
    \toprule
        \textbf{Method} & \textbf{Accept Rate (\%)} \\
    \toprule
        \textbf{Text+LLM} & 94.5\\
        \textbf{Whisper+LLM} & 91.3 \\
        \textbf{CTC+LLM} & 88.6 \\
        \textbf{ASR pretraining} & \phantom{0}0.0 \\
        \midrule
        \textbf{BLSP} & 88.5 \\
        \quad +RP & 88.3 \\
    \bottomrule
    \end{tabular}
    \caption{ChatGPT evaluation using acceptable rate.}
    \label{tab:qa}
\end{table}

\subsection{Analysis}
\paragraph{Effectiveness as a Pre-Training Strategy} 

\begin{table*}[htbp]
\centering
\scriptsize
\begin{tabular}{l|cccccccc}
    \toprule
    \textbf{Method} & \textbf{en-de} & \textbf{en-es} & \textbf{en-fr} & \textbf{en-it} & \textbf{en-nl} & \textbf{en-pt} & \textbf{en-ro} & \textbf{en-ru} \\
    \toprule
    \textbf{w/o pretraining} & 21.1 / 74.4 & 25.4 / 76.1 & 29.9 / 75.6 & 20.6 / 76.1 & 23.6 / 76.8 & 25.3 / 76.7 & 16.4 / 74.7 & 13.7 / 73.5 \\
    \textbf{ASR pretraining} & 22.7 / 76.6 & \textbf{27.9} / 78.7 & \textbf{32.1} / 77.7 & 22.3 / 78.2 & 25.4 / 78.7 & 27.3 / 79.6 & 18.6 / 77.4 & 14.9 / 76.2 \\
    \textbf{BLSP} & \textbf{23.3} / \textbf{77.7} & 27.4 / \textbf{79.5} & 31.9 / \textbf{78.5} & \textbf{23.2} / \textbf{79.0} & \textbf{26.4} / \textbf{80.0} & \textbf{28.5} / \textbf{80.4} & \textbf{19.2} / \textbf{78.6} & \textbf{15.6} / \textbf{77.3} \\
    \bottomrule
\end{tabular}
\caption{ST results (BLEU / COMET) of fine-tuned models on MUST-C.}
\label{tab:mustc_sup}
\end{table*}

We evaluate the BLSP method's effectiveness as a pre-training strategy for downstream tasks, with a focus on speech translation. To do this, we follow the same translation instruction as used in zero-shot translation tasks and fine-tune the primary BLSP model to predict target language translations directly from speech inputs. This fine-tuning process utilizes training data across eight language pairs from the MUST-C dataset. We apply LoRA (Low-Rank Adaptation) \citep{hu2021lora} to modify the key, query, value, and output layers of the LLM's self-attention mechanism, setting LoRA hyperparameters to $R=16$ and $\alpha=16$. We also update the speech encoder and the modality adapter's parameters to enhance model performance. For context, we compare these results with a commonly used pre-training approach, specifically ASR pre-training, as detailed in Section~\ref{sec:baselines}.

As illustrated in Table \ref{tab:mustc_sup}, our primary BLSP model exhibits an advantage in pre-training the modality adapter for the downstream speech translation task, achieving substantial improvements over random initialization. While pre-training the modality adapter using the ASR task proves beneficial, it may introduce a bias that hinders its ability to generalize across different downstream tasks. This limitation is highlighted by the superior performance of our BLSP approach over ASR pre-training. Our method achieves notably higher COMET scores across all translation directions and higher BLEU scores in six out of the eight directions evaluated.

\paragraph{Effectiveness in Speech-Text Alignment}

\begin{figure}[tbp]
    \centering
    \includegraphics[width=0.5\textwidth]{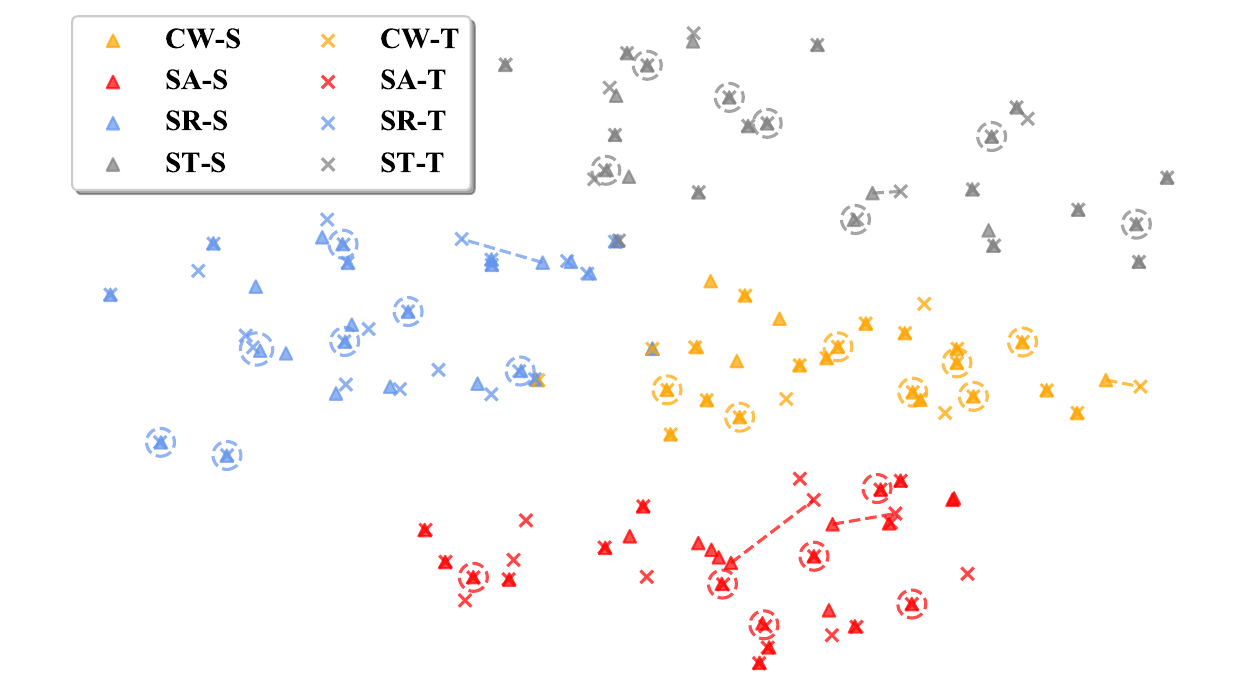}
    \caption{T-SNE visualization of feature representations learned from BLSP. Selected paired speech and text inputs are highlighted using dashed lines and circles.}
    \label{fig:continuation}
\end{figure}

We evaluate the effectiveness of the BLSP method in aligning speech and text inputs, using the procedures outlined in Section~\ref{sec:background}. As illustrated in Figure~\ref{fig:continuation}, the distribution of learned representations from speech inputs by the primary BLSP model no longer significantly differs from that of text inputs. This is a departure from the results observed with the ASR task, as depicted in Figure~\ref{fig:vis_asr}. The representations of speech inputs now share the same distribution as those of text inputs, with the representations of paired speech and text inputs being closely aligned, often overlapping. In Appendix \ref{app:cos}, we provide quantitative evidence that our BLSP model can generate distinct representations for the same speech input under different instructions, and that the representations for paired speech and text inputs closely match when given the same instructions. These results indicate that the BLSP approach effectively aligns speech and text inputs within the same space, thereby extending the instruction-following capabilities of LLMs to speech inputs.


\begin{figure}[tbp]
    \centering
    \includegraphics[width=0.45\textwidth]{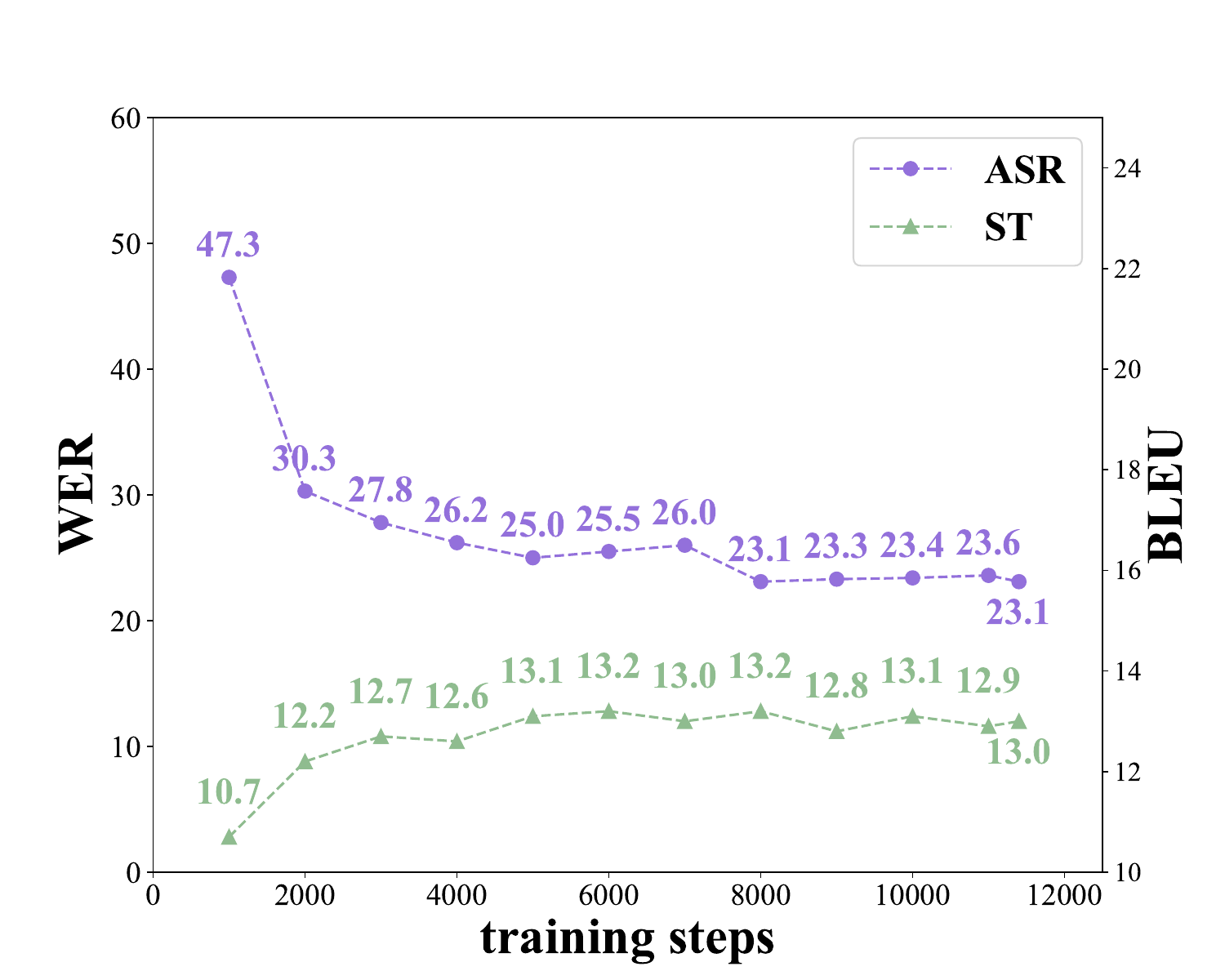}
    \caption{ASR and ST results at different training steps in one epoch.}
    \label{fig:scale}
\end{figure}

\paragraph{Impact of Data Size} We evaluate the impact of data size on model performance within the BLSP approach, utilizing measurements on out-of-domain datasets, specifically TED-LIUM 3 for zero-shot ASR performance and MUST-C en-de direction for zero-shot ST performance. In our experimental setup, we limit model training to a single epoch since the training loss converges well before the completion of one epoch. Consequently, we employ its performance at various training steps (approximately 0.8 million training samples for every 1,000 updates) as an estimate of its performance at different data scales. As shown in Figure \ref{fig:scale}, we observe rapid improvement in model performance during the early stages of training, 
followed by convergence after approximately 8,000 updates (equivalent to around 6 million training samples).

\begin{figure*}[htp]
    \centering
    \begin{minipage}[t]{0.45\textwidth}
    \centering
    \includegraphics[width=\textwidth]{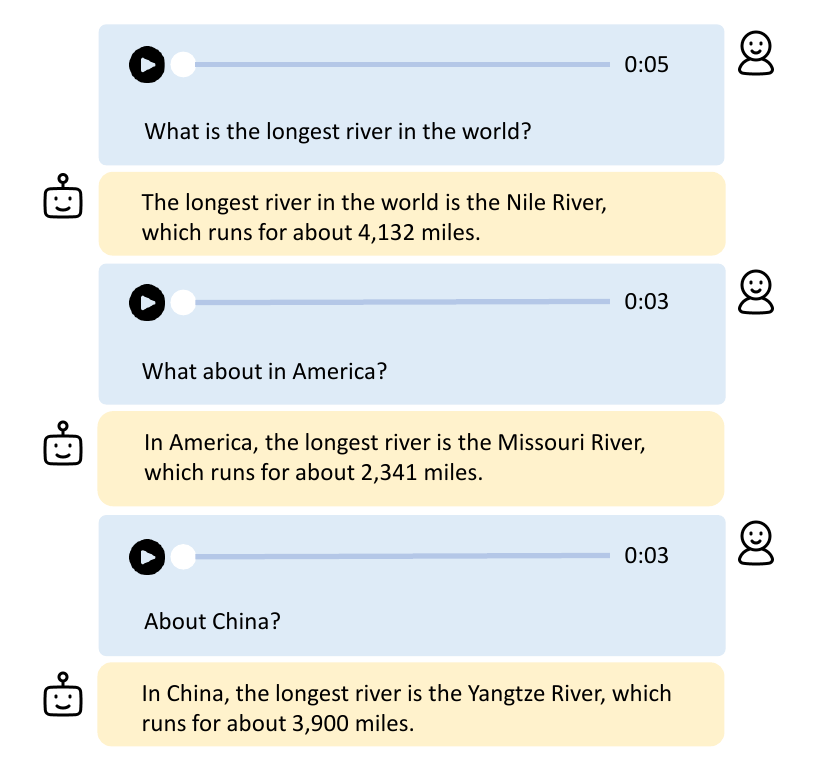}
    \caption{Speech conversation in English.}
    \label{fig:demo_english}
    \end{minipage}
    \begin{minipage}[t]{0.45\textwidth}
    \centering
    \includegraphics[width=\textwidth]{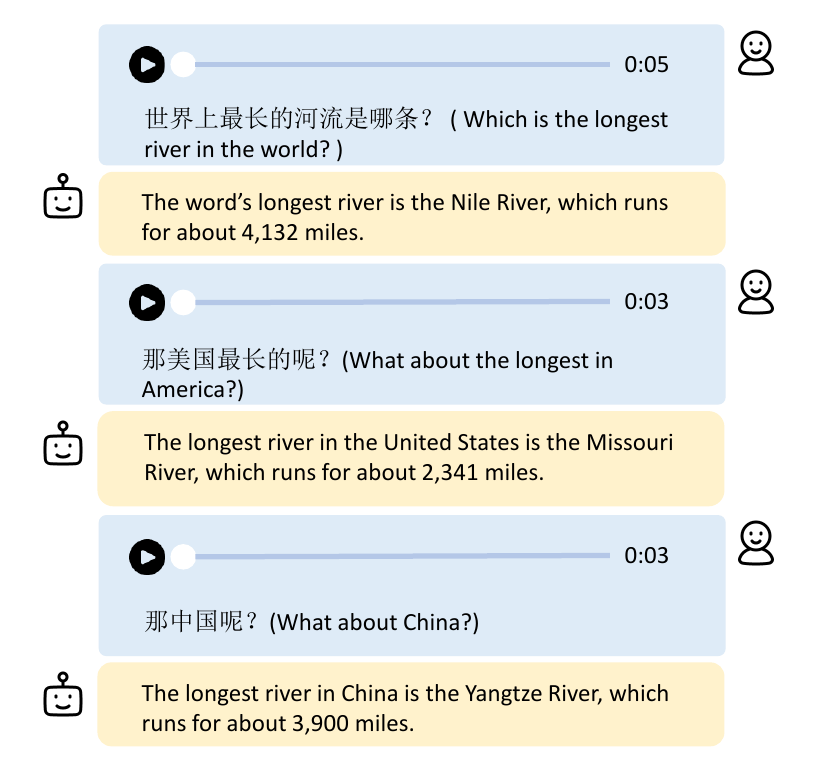}
    \caption{Speech conversation in Mandarin.}
    \label{fig:demo_chinese}
    \end{minipage}
\end{figure*}

\subsection{Cross-Modal Conversation}
We have observed that the BLSP approach can enable multi-turn conversation capabilities with LLMs using speech, thereby extending their remarkable conversational capabilities learned from text-only data to spoken languages. Figure \ref{fig:demo_english} illustrates an example of engaging in a spoken conversation in English with the model. More examples are presented in Appendix \ref{app:example}. Longer video demonstrations are available online\footnote{Video demos are available at \url{https://cwang621.github.io/blsp.github.io/}}.

\subsection{Emergence of Multilingual Capabilities}

Despite being trained solely on English ASR data for behavior alignment in continuation writing, we have observed that the BLSP model demonstrates an understanding of non-English speech inputs. This can be attributed to the multilingual capabilities of both the speech encoder (Whisper, \citet{radford2022robust}) and the LLM (Llama-2, \citet{touvron2023llama}), as well as the specific design of the BLSP training process. Note that both the speech encoder and the LLM remain frozen during BLSP training, suggesting that despite training solely on English data, the modality adapter can learn to project the multilingual space in Whisper encoder's output to the multilingual space for the LLM.

\begin{table}[tbp]
\centering
\footnotesize
\begin{tabular}{lcccc}
    \toprule
    \multirow{2}{*}{\textbf{Method}} & \multicolumn{1}{c}{\textbf{BSTC}} & \multicolumn{2}{c}{\textbf{MSLT}} \\ 
    \cmidrule(lr){2-2} \cmidrule(lr){3-4}
    & \textbf{zh-en} & \textbf{de-en} & \textbf{fr-en} \\ 
    \toprule
    \textbf{Text+LLM} & 16.1 / 58.7 & 32.8 / 84.2 & 29.6 / 76.3 \\
    \textbf{Whisper+LLM} & 11.1 / 54.3 & 25.3 / 79.4 & 24.1 / 71.5 \\
    \textbf{CTC+LLM} & \phantom{0}1.1 / 41.3  & \phantom{0}5.3 / 60.4 & \phantom{0}4.0 / 53.4 \\
    \textbf{BLSP} & \phantom{0}5.0 / 49.8 & 13.1 / 70.9 & 13.4 / 64.8 \\
    \bottomrule
\end{tabular}
\caption{ST results (BLEU / COMET) in X-to-English directions.}
\label{tab:multilingual}
\end{table}

To quantitatively measure the multilingual capabilities, we evaluate the speech translation performance of our BLSP model in the Chinese (zh) to English (en) direction on BSTC \citep{zhang2021bstc} and in the German (de) and French (fr) to English (en) directions on MSLT \citep{federmann2016microsoft}. As shown in Table \ref{tab:multilingual}, the BLSP model demonstrates reasonable multilingual translation competency for source languages that were not observed during behavior alignment training. We note that there is a significant gap in translation quality, as measured by both BLEU and COMET, when compared to Whisper+LLM and Text+LLM. This highlights the potential for further advancements in multilingual training. On the other hand, the cascaded model CTC+LLM, which was trained on English data, does not have cross-lingual capability.


As illustrated in Figure \ref{fig:demo_chinese}, our model is capable of engaging in multi-turn conversations with non-speech (Mandarin) speech input. It is worth mentioning that the model's responses are always in English. This is a direct result of the English-only training procedure in BLSP, where the continuations are consistently in English. This observation also suggests that there is benefit in incorporating multilingual training in behavior alignment for future research.

\section{Related Works}
Due to the lack of space, please see Appendix \ref{app:related_works} for a discussion on related works.


\section{Conclusion}
In this paper, we introduce the BLSP approach, which bootstraps language-speech pre-training through behavior alignment. Our training procedure is straightforward, requiring only learning of a lightweight modality adapter through a novel utilization of speech recognition training data. As evidenced by quantitative evaluations in speech recognition, speech translation, spoken language understanding, and illustrated through multi-turn conversation demonstrations, BLSP effectively extends the remarkable language capabilities of LLMs to speech, enabling direct interaction with LLMs using speech input. 
BLSP represents a fresh and valuable perspective for achieving cross-modal alignment in LLMs, and there are numerous directions for expansion and improvement in future research. 

\section*{Limitations}

Although our BLSP approach can extend the remarkable language capabilities of LLMs to speech, as evidenced by quantitative evaluations and illustrative demonstrations, there are several limitations in our current study.

\paragraph{Alignment Quality.} As indicated by the quantitative evaluations, there exists a substantial performance gap when using speech input as opposed to the cascaded approach. Our approach to behavior alignment of continuation writing, in its current form, tends to align speech and text at a semantic level that restricts its capacity to capture detailed phonetic information. Exploring more fine-grained loss designs or approaches for constructing more fine-grained training data, including in combination with speech recognition, speech translation, or general speech instruction data, is worthy of further investigation.

\paragraph{Paralinguistic Information.} In this study, we mainly focus on aligning speech and text in the semantic space, without addressing the paralinguistic aspects of spoken language that cannot be simply described by words, such as emotions, tones, and intentions. It is possible to capture and incorporate paralinguistic information with LLMs by leveraging data from more diverse speech-related tasks, such as speaker identification, keyword spotting, and speech emotion recognition.

\paragraph{Safety and Ethics.} The use of continuous speech representations in our BLSP model could make it more susceptible to adversarial attacks and can potentially compromise the LLM's established adherence to the HHN criteria (Harmless, Helpful, Honest). This is an area that is worthy of future research, both in identifying weaknesses and searching for solutions.

\paragraph{Broader Applicability.} While our study has focused on the behavior alignment of continuation writing for speech-text alignment, the fundamental principles underlying this approach could have broader applicability. This involves expanding existing paired data in creative ways with the assistance of LLMs, ultimately benefiting LLMs. We leave it to future studies to extend this approach to diverse scenarios, including vision-language and multilingual alignments.

\bibliography{custom}

\appendix

\section{Implementation Details of ASR Pretraining}
\label{app:background}

In the pre-experiments, the model architecture and training dataset used are the same as in BLSP. The only difference is that in BLSP, the modality adapter is trained using the continuation task, while in the pre-experiments, the modality adapter is trained to predict the ground-truth transcript.
Additionally, similar to the approach in SpeechGPT \citep{zhang2023speechgpt}, we utilize GPT-4 to generate 100 distinct text instructions for prompting ASR tasks. These instructions are concatenated before the speech input.
For visualization purposes, we construct cross-modal prompts to extract features using the instructions shown in Table \ref{tab:ins_background}. We then apply t-SNE dimensionality reduction mapping to all samples from the LibriSpeech test set.


\begin{table}[htbp]
    \centering
    \footnotesize
    \begin{tabular}{p{0.45\textwidth}}
    \toprule
    \textbf{CW:} \emph{Please continue the following sentence.} \\
    \midrule
    \textbf{SA:} \emph{Please classify the emotional tone of the following text.} \\
    \midrule
    \textbf{SR:} \emph{Please transcribe the following audio into English text.} \\
    \midrule
    \textbf{ST:} \emph{Please translate the following English text into German text.} \\
    \bottomrule
    \end{tabular}
\caption{Instructions used for extracting cross-modal representations.}
\label{tab:ins_background}
\end{table}

To further demonstrate the overfitting problem in the ASR task, as illustrated in Figure \ref{fig:vis_asr}, we present the average cosine similarity between the learned representations of the same input across different task instructions in Table \ref{tab:cos_asr_1}. Notably, the representations for speech input are remarkably similar regardless of the task instruction used, indicating a deficiency in following instructions. Additionally, Table \ref{tab:cos_asr_2} highlights consistently low similarity scores between paired speech and text input representations under the same task instructions, suggesting a lack of alignment between the representations of speech and text inputs.


\begin{table}[htbp]
    \centering
    \begin{tabular}{c|cccc}
    \toprule
        & \textbf{CW-S} & \textbf{SA-S} & \textbf{SR-S} & \textbf{ST-S} \\
    \toprule
        \textbf{CW-S} &  1.000 &  0.997 &  0.997 &  0.991 \\
        \textbf{SA-S} &  0.997 &  1.000 &  0.997 & 0.992 \\
        \textbf{SR-S} &  0.997 & 0.997 &  1.000 &  0.993 \\
        \textbf{ST-S} &  0.991 &  0.992 &  0.993 &  1.000 \\
    \bottomrule
    \end{tabular}
    \caption{Average similarity between representations of the same speech inputs under different task instructions learned from ASR task.}
    \label{tab:cos_asr_1}
\end{table}

\begin{table}[htbp]
    \centering
    \begin{tabular}{cccc}
    \toprule
        \textbf{CW} & \textbf{SA} & \textbf{SR} & \textbf{ST} \\
    \toprule
         0.270 &  0.106 &  0.328 &  0.176 \\
    \bottomrule
    \end{tabular}
    \caption{Average similarity between representations of paired speech/text inputs under the same task instructions learned from ASR task.}
    \label{tab:cos_asr_2}
\end{table}

\begin{table*}[tbp]
\centering
\footnotesize
\begin{tabular}{l|ccccc}
    \toprule
    \textbf{Method} & \textbf{en-ca} & \textbf{en-de} & \textbf{en-id} & \textbf{en-sl} & \textbf{en-sv} \\
    \midrule
    \textbf{Text+LLM} & 24.8  & 21.9  & 22.0  & 12.9  & 27.8 \\
    \textbf{Whisper+LLM} & 19.2  & 17.3  & 15.8  & 10.0  & 22.2  \\
    \textbf{CTC+LLM} & 14.5 & 14.0 & 12.3 & \phantom{0}7.9 & 17.5 \\
    \midrule
    \textbf{BLSP} & 13.9  & 14.0  & 12.1  & \phantom{0}7.0  & 16.6  \\
    \quad + RP & 14.8 & 14.9 & 12.3 & \phantom{0}7.9 & 19.3 \\
    \bottomrule
\end{tabular}
\caption{ST results on in-domain dataset CoVoST-2.}
\label{tab:covost}
\end{table*}

\begin{table*}[tbp]
\centering
\footnotesize
\begin{tabular}{l|cccccccc}
    \toprule
    \textbf{Method} & \textbf{en-de} & \textbf{en-es} & \textbf{en-fr} & \textbf{en-it} & \textbf{en-nl} & \textbf{en-pt} & \textbf{en-ro} & \textbf{en-ru} \\
    \midrule
    \textbf{Text+LLM} & 20.2  & 21.7  & 24.2  & 20.5  & 24.4  & 20.3 & 13.3 & 13.3 \\
    \textbf{ASR+LLM} & 16.9 & 19.2 & 20.5 & 16.4 & 20.7  & 17.1  & 10.9  & 11.4  \\
    \textbf{CTC+LLM} & 13.6 & 15.3 & 15.7 & 12.6 & 17.0 & 13.9 & \phantom{0}8.2 & \phantom{0}9.9 \\
    \midrule
    \textbf{BLSP} & 13.0  & 14.7  & 14.4  & 11.6  & 17.0  & 12.6  & \phantom{0}6.2  & \phantom{0}9.2 \\
    \quad +RP & 14.2 & 16.4 & 17.5 & 13.2 & 17.5 & 16.1 & \phantom{0}9.2 & 10.2 \\
     \bottomrule
\end{tabular}
\caption{ST results on out-of-domain dataset MUST-C.}
\label{tab:mustc}
\end{table*}

\section{Detailed Results for Zero-Shot Speech-to-Text Tasks}
\label{app:results}

We present the performance of speech translation in each direction, as illustrated in Tables \ref{tab:covost} and Table \ref{tab:mustc}. 
For CoVoST-2, we evaluate our method on five translation directions: English (en) to Catalan (ca), German (de), Indonesian (id), Slovenian (sl), and Swedish (sv).
Additionally, we conduct experiments on MUST-C for all eight translation directions: English (en) to German (de), Spanish (es), French (fr), Italian (it), Dutch (nl), Portuguese (pt), Romanian (ro), and Russian (ru). 
The instructions used for each speech-to-text generation task are presented in Table \ref{tab:ins}.

\begin{table}[htbp]
    \centering
    \footnotesize
    \begin{tabular}{p{0.45\textwidth}}
    \toprule
    \textbf{ASR:} \emph{Please repeat the following words.} \\
    \midrule
    \textbf{ST:} \emph{Please translate the following English text into <target> text.} \\
    \midrule
    \textbf{SNIPS:} \emph{Please classify the intent of the text, choose from [DecreaseBrightness, IncreaseBrightness, SetLightBrightness, SetLightColor, SwitchLightOff, SwitchLightOn].} \\
    \midrule
    \textbf{FSC:} \emph{Please classify the intent of the text, choose from [bring newspaper, deactivate lamp, change language English, deactivate music, increase heat, change language Korean, change language none, bring shoes, change language German, activate lights, bring socks, change language Chinese, decrease heat, decrease volume, increase volume, activate music, activate lamp, bring juice].} \\
    \midrule
    \textbf{SLUE-VoxCeleb:} \emph{Please classify the emotional tone of the text as either positive, negative, or neutral.} \\
    \bottomrule
    \end{tabular}
\caption{Instructions used for speech-to-text generation tasks.}
\label{tab:ins}
\end{table}

\section{The Data Construction and Evaluation Process of General-Purpose QA}
\label{app:qa}

For our evaluation on the general-purpose question answering task, we selected 1460 speech-text pairs from the GigaSpeech test set. The selected texts contain 40-60 words to ensure that the samples encompass relatively complete semantics.
To formulate questions based on these transcripts, we utilized ChatGPT. As shown in Listing \ref{lst:data}, we provided ChatGPT with the transcript as the input, and the task for ChatGPT was to generate a suitable question based on the given text input.


In the next step, we used different models to generate responses to the questions posed by ChatGPT in the previous step. We then employed ChatGPT again to evaluate the acceptability of these generated answers.
As shown in Listing \ref{lst:eval}, we provided ChatGPT with the question, the ground-truth transcript, and the answer generated by different models. The task for ChatGPT was to evaluate whether the generated answer is acceptable.


\section{Quantitative Analysis of Representations from BLSP}
\label{app:cos}

As depicted in Table \ref{tab:cos_blsp_1}, the representations of speech inputs learned from BLSP are distinct under various task instructions, unlike in Table \ref{tab:cos_asr_1} for ASR task. Table \ref{tab:cos_blsp_2} illustrates the average cosine similarity between representations of paired speech and text inputs learned from BLSP, revealing a high level of similarity between the two, as opposed to the low similarity depicted in Table \ref{tab:cos_asr_2} for ASR task. We want to point out that there remains a notable gap between the representations for speech and text inputs that is worthy of future research.

\begin{table}[htbp]
    \centering
    \begin{tabular}{c|cccc}
    \toprule
        & \textbf{CW-S} & \textbf{SA-S} & \textbf{SR-S} & \textbf{ST-S} \\
    \midrule
        \textbf{CW-S} &  1.000 & 0.494 & 0.745 & 0.381 \\
        \textbf{SA-S} &  0.494 & 1.000 & 0.501 & 0.278 \\
        \textbf{SR-S} & 0.745 & 0.501 & 1.000 & 0.477 \\
        \textbf{ST-S} & 0.381 & 0.278 & 0.477 & 1.000 \\
    \bottomrule
    \end{tabular}
    \caption{Average similarity between representations of the same speech inputs under different task instructions learned from BSLP.}
    \label{tab:cos_blsp_1}
\end{table}

\begin{table}[htbp]
    \centering
    \begin{tabular}{cccc}
    \toprule
        \textbf{CW} & \textbf{SA} & \textbf{SR} & \textbf{ST} \\
    \midrule
        0.785 & 0.866 & 0.808 & 0.900 \\
    \bottomrule
    \end{tabular}
    \caption{Average similarity between representations of paired speech/text inputs under the same task instructions learned from BLSP.}
    \label{tab:cos_blsp_2}
\end{table}

\section{Selected Examples of Cross-Modal Conversation}
\label{app:example}

\begin{figure*}[htp]
    \centering
    \begin{minipage}[t]{0.48\textwidth}
    \centering
    \includegraphics[width=\textwidth]{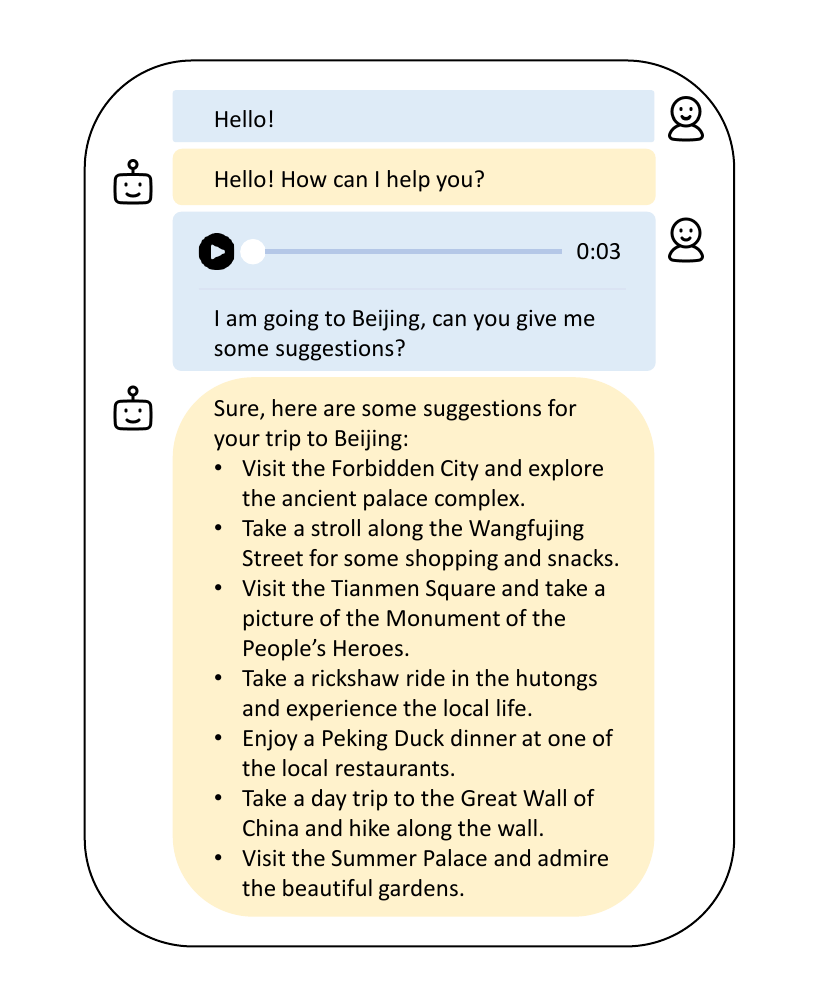}
    \end{minipage}
    \begin{minipage}[t]{0.48\textwidth}
    \centering
    \includegraphics[width=\textwidth]{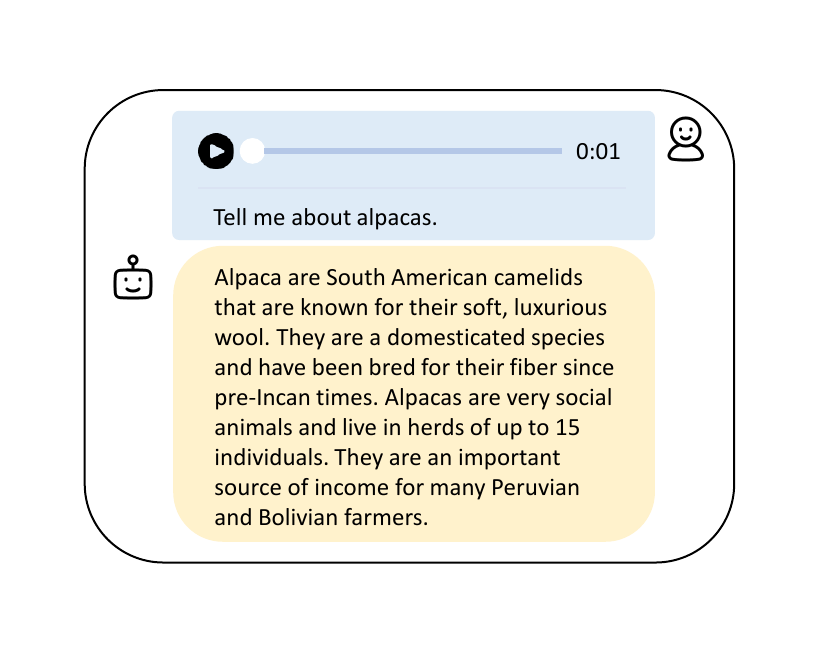}
    \end{minipage}
    \begin{minipage}[t]{0.48\textwidth}
    \centering
    \includegraphics[width=\textwidth]{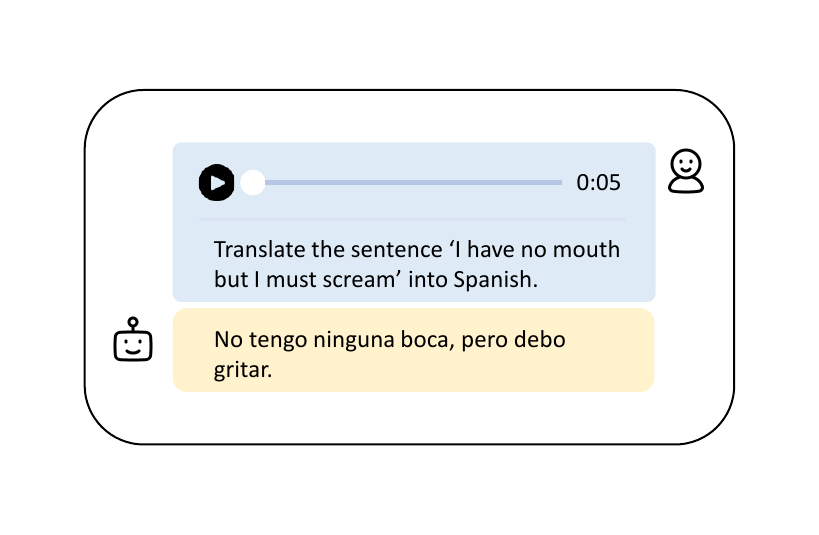}
    \end{minipage}
    \begin{minipage}[t]{0.48\textwidth}
    \centering
    \includegraphics[width=\textwidth]{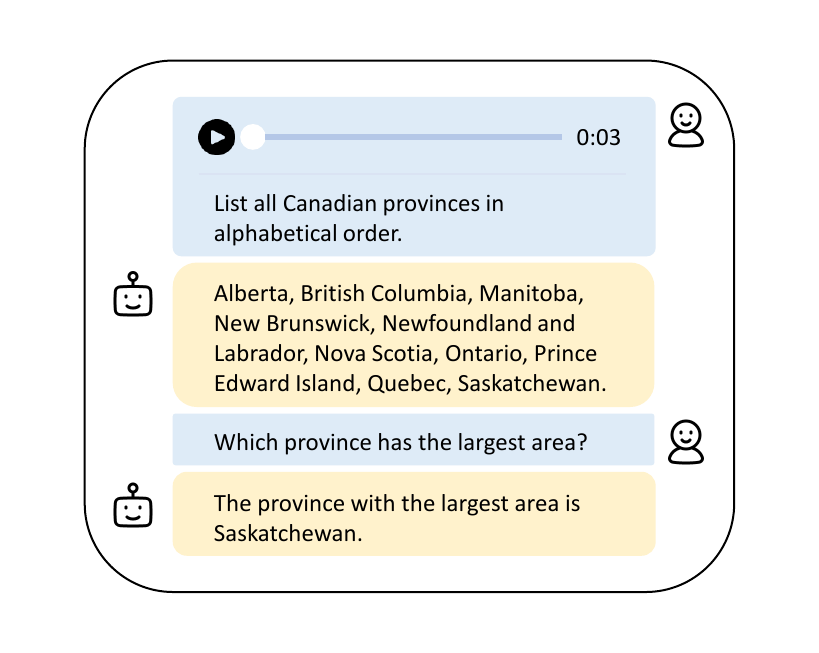}
    \end{minipage}
    \caption{Selected examples of cross-modal conversation using BLSP model.}
    \label{fig:more_demos}
\end{figure*}

As demonstrated in Figure \ref{fig:more_demos}, BLSP provides expanded mechanisms to interact with LLMs. Users can freely switch between text and speech inputs, and directly employ speech instructions to carry out speech-to-text tasks.

\begin{figure*}[htbp]
\small
\begin{lstlisting}[caption={The prompt used to generate general-purpose QA data.},label={lst:data},basicstyle=\ttfamily,breaklines=true,language=Tex,breakindent=0pt]
Please ask a question about the input and then answer the question based on the input. The output format should be in json and contains question and the response.
Example:
input: ah yeah good day and welcome to this instructional video on how to ah wash your car um with a baby . basically , you just ask them to do it . you know they love this kind of stuff this bubbles and a brush and .
output: {"question": What is the video about?, "answer": the video is about how to wash your car um with a baby.}
input: it is the gibraltar strait where you lost control and then you dived down ... one of those cases where you let the wings go in the clouds but you lose orientation completely
output: {"question": Where did the incident occur?, "answer": Gibraltar Strait.}
BEGIN:
input: ${transcript}
output:
\end{lstlisting}
\end{figure*}

\begin{figure*}[htbp]
\small
\begin{lstlisting}[caption={The prompt used to evaluate whether a response is acceptable.},label={lst:eval},basicstyle=\ttfamily,breaklines=true,language=Tex,breakindent=0pt]
Given a question, related input, and answer, please help determine whether the answer is acceptable.
The output choose from acceptable or unacceptable.
Question: ${question}
Input: ${transcript}
Answer: ${answer}
Your output: 
\end{lstlisting}
\end{figure*}

\section{Related Works}
\label{app:related_works}

\paragraph{Multi-Modal Large Language Models} Current multi-modal large language models have been prominently focusing more on visual modality \citep{openai2023gpt,yin2023survey}. These models utilize a pre-trained visual encoder to extract key visual features from images, which are then combined with text inputs to generate relevant outputs. PaLM-E \citep{driess2023palm} combines the huge 540B PaLM \citep{chowdhery2022palm} and the 22B Vision Transformer (ViT) \citep{dosovitskiy2020image} to create the largest vision-language model currently reported. Since it would be costly to train a large multi-modal model in an end-to-end manner, many works introduce a learnable interface between the pre-trained visual encoder and LLM to connect different modalities while freezing the
parameters of the pre-trained models. Flamingo \citep{alayrac2022flamingo}, BLIP-2 \citep{li2023blip} and X-LLM \citep{chen2023x} leverage a group of learnable query tokens to extract information in a query-based manner. LLaVa \citep{liu2023visual} connects the pre-trained CLIP \citep{radford2021learning} encoder and Vicuna \citep{vicuna2023} with a simple projection layer. LLaMA-Adapter \citep{gao2023llama} and LaVIN \citep{luo2023cheap} explore a parameter-efficient tuning manner, introducing a lightweight adapter module during training. Recent research has extended the above-mentioned approach to ``audio'' \citep{gong2023listen}, which refers to natural sound, such as thunder and chirp. However, there is still a lack of exploration when it comes to human speech.

\paragraph{Interact with LLMs through Speech} After the introduction of ChatGPT,  several studies have focused on combining specialized speech models with LLMs, allowing for speech interaction with these language models. Initial endeavors in this field (e.g., HuggingGPT \citep{shen2023hugginggpt}, AudioGPT \citep{huang2023audiogpt}) employed a cascading model structure, linking LLMs with additional ASR and TTS models to enable speech input and output. These models showcase heightened intricacy, require substantial resources, and are susceptible to the inevitable issue of error accumulation. Recent works have started to explore end-to-end model architectures. SpeechGPT \citep{zhang2023speechgpt} takes the discretized output of a speech model in self-supervised training and treats it as a specialized linguistic unit, training it alongside a large language model. However, due to the high sampling frequency of the discrete unit, it is difficult for this method to achieve multiple rounds of dialogue. LLaSM \citep{shu2023llasm} has constructed an extensive speech instruction dataset intended for training the modal adapter to attain modality alignment. Their methodology is predominantly data-driven, with a lesser emphasis on the explicit design of modality alignment. 

\end{document}